%% file: ARXIV_2022_Object_Permanence.tex
\documentclass[a4paper,conference]{IEEEtran}

\usepackage{graphicx}
\usepackage{amsmath}
\usepackage{amssymb}
\usepackage{booktabs}
\usepackage[table,xcdraw]{xcolor}  
\usepackage{multirow}
\usepackage{xcolor}
\usepackage[pagebackref,breaklinks,colorlinks]{hyperref}
\usepackage[normalem]{ulem}
\useunder{\uline}{\ul}{}

\usepackage[capitalize]{cleveref}
\crefname{section}{Sec.}{Secs.}
\Crefname{section}{Section}{Sections}
\Crefname{table}{Table}{Tables}
\crefname{table}{Tab.}{Tabs.}

\newcommand{\ie}{i.e. }
\newcommand{\eg}{e.g. }

\newcommand{\todo}[1]{}
\renewcommand{\todo}[1]{\textcolor{red}{@TODO: {#1}}}
\newcommand{\todoMF}[1]{}
\renewcommand{\todoMF}[1]{\textcolor{orange}{@TODO MF: {#1}}}
\newcommand{\todoStudent}[1]{}
\renewcommand{\todoStudent}[1]{\textcolor{cyan}{@TODO Student: {#1}}}

\newcommand{\B}[1]{}
\renewcommand{\B}[1]{\textbf{{#1}}}

\newcommand{\lorem}[1]{}
\renewcommand{\lorem}[1]{\todo{REMOVE}~\textcolor{gray}{Lorem ipsum dolor sit amet, consetetur sadipscing elitr, sed diam nonumy eirmod tempor invidunt ut labore et dolore magna aliquyam erat, sed diam voluptua. At vero eos et accusam et justo duo dolores et ea rebum. Stet clita kasd gubergren, no sea takimata sanctus est Lorem ipsum dolor sit amet. Lorem ipsum dolor sit amet, consetetur sadipscing elitr, sed diam nonumy eirmod tempor invidunt ut labore et dolore magna aliquyam erat, sed diam voluptua. At vero eos et accusam et justo duo dolores et ea rebum. Stet clita kasd gubergren, no sea takimata sanctus est Lorem ipsum dolor sit amet.
		Lorem ipsum dolor sit amet, consetetur sadipscing elitr, sed diam nonumy eirmod tempor invidunt ut labore et dolore magna aliquyam erat, sed diam voluptua. At vero eos et accusam et justo duo dolores et ea rebum. Stet clita kasd gubergren, no sea takimata sanctus est Lorem ipsum dolor sit amet. Lorem ipsum dolor sit amet, consetetur sadipscing elitr, sed diam nonumy eirmod tempor invidunt ut labore et dolore magna aliquyam erat, sed diam voluptua. At vero eos et accusam et justo duo dolores et ea rebum. Stet clita kasd gubergren, no sea takimata sanctus est Lorem ipsum dolor sit amet.}}

\newcommand{\loremSmall}[1]{}
\renewcommand{\loremSmall}[1]{\todo{REMOVE}~\textcolor{gray}{Lorem ipsum dolor sit amet, consetetur sadipscing elitr, sed diam nonumy eirmod tempor invidunt ut labore et dolore magna aliquyam erat, sed diam voluptua. At vero eos et accusam et justo duo dolores et ea rebum. Stet clita kasd gubergren, no sea takimata sanctus est Lorem ipsum dolor sit amet. Lorem ipsum dolor sit amet, consetetur sadipscing elitr, sed diam nonumy eirmod tempor invidunt ut labore et dolore magna aliquyam erat, sed diam voluptua. At vero eos et accusam et justo duo dolores et ea rebum.}}

\begin{document}
	\title{Object Permanence in Object Detection\\ Leveraging Temporal Priors at Inference Time}
	
	\author{
		\IEEEauthorblockN{
			Michael Fürst\IEEEauthorrefmark{1}\IEEEauthorrefmark{2},
			Priyash Bhugra\IEEEauthorrefmark{2},
			René Schuster\IEEEauthorrefmark{1} and
			Didier Stricker\IEEEauthorrefmark{1}\IEEEauthorrefmark{2}}
		\IEEEauthorblockA{\IEEEauthorrefmark{1}DFKI - German Research Center for Artificial Intelligence\\
			Trippstadter Strasse 122, 67663 Kaiserslautern, Germany\\
			{\tt\small firsname.lastname@dfki.de}}
		\IEEEauthorblockA{\IEEEauthorrefmark{2}Technische Universität Kaiserslautern\\
			67663 Kaiserslautern, Germany\\
		}
	}
	
	\maketitle
	\input{content/0_abstract}
	\IEEEpeerreviewmaketitle
	
	\input{content/1_introduction}

	\input{content/2_related_work}
	\input{content/3_approach}
	\input{content/4_results}

	\input{content/5_conclusions}
	\input{content/6_acknowledgment}

	{\small
		\bibliographystyle{ieee_fullname}
		\bibliography{egbib}
	}
\end{document}

%% file: content/0_abstract.tex
\begin{abstract}
Object permanence is the concept that objects do not suddenly disappear in the physical world.
Humans understand this concept at young ages and know that another person is still there, even though it is temporarily occluded.
Neural networks currently often struggle with this challenge.
Thus, we introduce explicit object permanence into two stage detection approaches drawing inspiration from particle filters.
At the core, our detector uses the predictions of previous frames as additional proposals for the current one at inference time.
Experiments confirm the feedback loop improving detection performance by a up to 10.3 mAP with little computational overhead.

Our approach is suited to extend two-stage detectors for stabilized and reliable detections even under heavy occlusion.
Additionally, the ability to apply our method without retraining an existing model promises wide application in real-world tasks.

%
%

\end{abstract}


%% file: content/1_introduction.tex
\section{Introduction}
Object detection has made significant advancements in the last decade and is applied to numerous new domains and live systems with increasing risks for humans and the environment.
However, operating on single frames detectors lack understanding of object permanence.
Object permanence~\cite{piaget2013construction} is the concept that objects in a physical world continue to exist despite the observers inability to sense them.
This can result in temporally unstable detections and poor occlusion robustness.

These approaches without an understanding of object permanence are applied in domains where safety is critical such as robotics and autonomous vehicles.
However, especially in these domains object permanence can greatly increase safety.
For example, when a pedestrian disappears behind a pole, in the physical world the object is still there, but a common perception algorithm without object permanence does not detect the person anymore.
Based on the most recent detections, the vehicle might optimize its trajectory to collide with the occluded pedestrians trajectory.
Without prior knowledge of preceding time steps the model is unable to recover a detection precisely if the features computed from a single frame are ambiguous.

\begin{figure}
	\centering
	\includegraphics[width=0.48\textwidth]{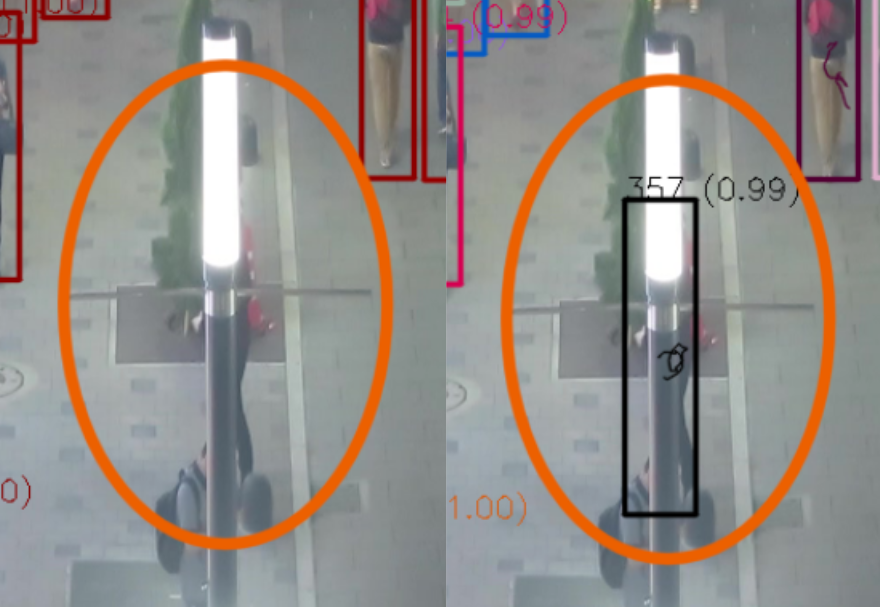}
	\caption{A single frame object detector (left) cannot detect the pedestrian behind the street light, our approach leveraging object permanence (right) detects the person despite the major occlusion lasting 30 frames avoiding the overhead of tracking approaches.}
	\label{fig:title}
\end{figure}

Object permanence is the understanding that an object still exists despite the inability to sense the object directly.
This knowledge can help with disambiguation of features in the case of occlusions.
In Figure~\ref{fig:title} a person behind the pole is imperceptible to a Faster-RCNN~\cite{fasterrcnn}.
When applying our approach giving the Faster-RCNN prior knowledge of the existence of the object, it is able to detect the person again without changing the weights of the model.

In tracking, object permanence is used in integrated trackers to generate tracklets where an object has been.
However, these trackers are computationally more expensive than detectors and require an altered training schema and special  sequential data, limiting their availability to much rarer tracking data sets.
Non integrated approaches like Kalman Filter~\cite{kalman1960new} can be applied on detectors without special training requirements.
However, temporal information is used only after the network.
Thus the network lacks temporal information, leading to missing detections and lower precision.

We present an approach to integrate object permanence in two-stage object detectors using dynamic proposal priors.
In contrast to full tracking approaches our goal is to improve detection performance with as little overhead and modification to the original task as possible.
Thus, our approach does not produce full tracklets, but has little computational overhead.
Further distinguishing our approach from current tracking approaches is that we integrate it into the model at test time, without re-training an existing two-stage single frame detector.

%% file: content/2_related_work.tex
\section{Related work}

\textbf{Object Detection} is the task of predicting a bounding box for an object in an image or scene.
The most common setup is 2D object detection, where the 2D axis aligned bounding box with a position and a size has to be predicted from an image.
To achieve this goal various solutions have been proposed which can be generally categorized into single stage approaches and two stage approaches.

Based on the work of LeCun et al.~\cite{lecun89, lecun98}, OverFeat~\cite{sermanet13} was one of the first single stage approaches for object detection, followed by SSD~\cite{SSD, fu17}, YOLO~\cite{yolo, yolov2, yolov3}, RetinaNet~\cite{retinanet} and others like the CenterNet~\cite{zhou2019objects}. These approaches consist of a single CNN as a feature extractor and then directly predict the positions and classes of the objects.

\begin{figure}
	\centering
	\includegraphics[width=0.48\textwidth]{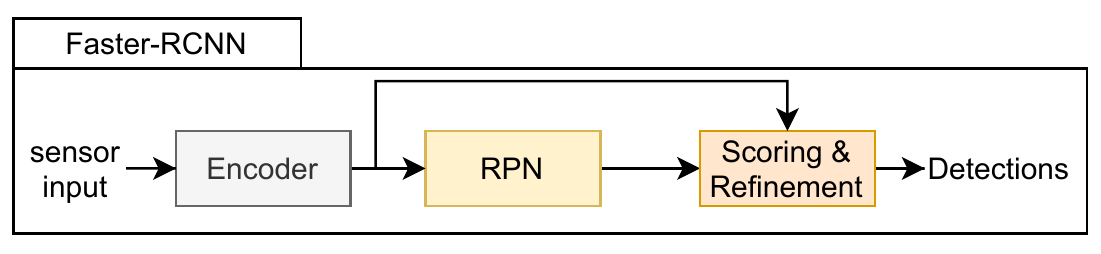}
	\caption{Faster-RCNN~\cite{fasterrcnn} consists of three main building blocks. First is a shared encoder which produces feature maps used by both stages. Next is the Region Proposal Network (RPN) generating the proposals of the first stage in Faster-RCNN. Last is the scoring and refinement of the proposals based on the features extracted from the feature map.}
	\label{fig:faster_rcnn}
\end{figure}

In contrast to these approaches, two-stage detectors are following the schema of extracting regions of interest (ROIs) and then classify each ROI.
Region-CNN~\cite{rcnn} uses selective search generate proposals.
Then, each ROI is classified by a CNN with a support vector machine (SVM).
However, as the inference speed was very slow, Fast-RCNN~\cite{fastrcnn} and Faster-RCNN~\cite{fasterrcnn} improved this by computing proposals using a CNN and sharing the feature encoder between the proposal stage named Region Proposal Network (RPN) and the classification and refinement stage often referred to as the second stage (Figure~\ref{fig:faster_rcnn}).
MRCNN~\cite{burlina2016mrcnn} replaces the slow selective search with a fast multi-target tracker to generate proposals in consecutive frames.
Later Mask-RCNN~\cite{he17} and others~\cite{he15, lin17b} were introduced.
The two stage-approaches can be further found in 3D object detectors~\cite{avod} and 3D pose estimation~\cite{furst2021hperl} where fusion of multiple sensor streams is done.

\textbf{Tracking-by-detection} are algorithms which build on above object detectors and extend them by a tracking module.
The core idea of these approaches is to use the detections, \ie the bounding boxes, and track objects based on them.
One of the simplest solutions is to use a Kalman Filter~\cite{kalman1960new, welch1995introduction} or a Particle Filter~\cite{boers2001particle} on the detections.

A particle filter is a three step process, using a large number of particles representing a multimodal distribution. The particles are scored in a measurement step and then resampled based on the scores to better represent the distribution.
Finally, it predicts the motion of particles over time.
In object tracking, a particle usually is a tracklet with a history of past positions, a current position and a size for the bounding box.

Over time more complex approaches like~\cite{bewley2016simple, fang2018recurrent, leal2016learning, schulter2017deep, xu2019spatial, zhu2018online} have evolved.
Tackling the association of detections and tracklets, methods using appearance features~\cite{wojke2017simple}, re-identification~\cite{tang2017multiple} and 3D shape information~\cite{sharma2018beyond} have been applied.
Improving occlusion cases with object permanence, \cite{shamsian2020learning} uses Faster-RCNN for initial detection and LSTMs for soft attention weights and refinement.

Known limitations of tracking-by-detection are discarding image information in the data association step or using expensive feature extractors as pointed out by CenterTrack~\cite{zhou2020tracking}.
Beyond that, the tracking and detection are separated and prior knowledge of previous frames in the tracker cannot be used for better detection.

\textbf{Integrated Detection and Tracking:} Due to the above limitations of tracking-by-detection effort has been spend on integrated detection and tracking.
Integrated Detection~\cite{zhang2018integrated} conditions the RPN and second stage on the tracklets.
Kang et al.~\cite{kang2017object, kang2017t} and Zhu et al.~\cite{zhu2017flow} use detection for a whole video segment and flow-warped intermediate features with a Faster-RCNN.
TransCenter~\cite{xu2021transcenter} uses transformers for tracking, while FairMOT~\cite{zhang2020fairmot} focuses on providing fair features for detection and re-identification to boost the performance of integrated tracking.

Tracktor~\cite{bergmann2019tracking} uses the second stage of Faster-RCNN to realign the boxes for the next time step and uses proposals form the RPN which have no substantial IoU with the existing tracks to generate new tracks.
This approach is most similar to ours, but using only proposals with a low IoU with existing tracks, they limit the potential for exploration and multi hypothesis modeling in case of pedestrian to pedestrian occlusions, which cause high IoUs while still representing different objects.

The box and IoU centric nature of Faster-RCNN based approaches is, according to CenterTrack~\cite{zhou2020tracking}, the cause for association difficulties.
Thus, they follow the idea of CenterNet~\cite{zhou2019objects} to predict the center of the bounding box as a heatmap and extends this by using the heatmap of the previous frame as an input to the current frame.
Further they predict the offset of the center from the current to the last frame to solve the association problem.
We agree, that IoU based suppression of detections is an issue for multi-modal bounding box distributions, specifically in the case of occlusions, but we emphasize that Faster-RCNN does not rely on the IoU based suppression to generate new proposals.
PermaTrack~\cite{tokmakov2021learning}~extends CenterTrack to N frames and explicit occlusion modeling during training to improve object permanence.

Other methods than IoU based suppression and selection have been successfully applied.
A particle filter has various particles which not directly suppress each other, but by scoring and re-sampling a selection process is taking place.
This makes particle filters optimal filters for complex multi-modal distributions, given sufficient particles.

In contrast to other approaches like Tracktor which used manual proposal selection via IoU thresholding, we propose an approach inspired by the implicit scoring of particle filters and integrate it into a pre-trained Faster-RCNN at inference time.
To our knowledge there have been no experiments on full integration of particle filter concepts and a Faster-RCNN.


%% file: content/3_approach.tex
\begin{figure*}
	\centering
	\includegraphics[width=0.98\textwidth]{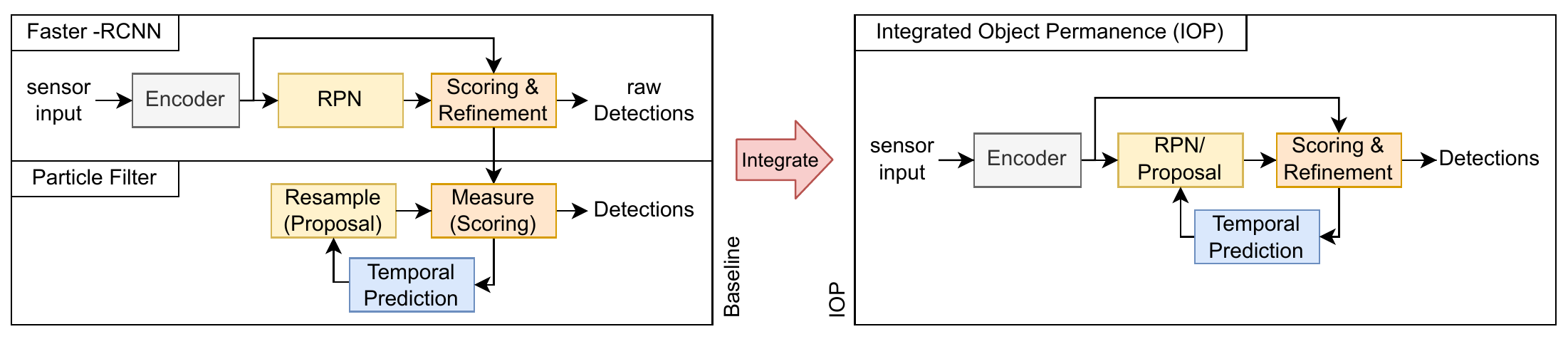}
	\caption{
		The core of our approach is integrating the particle filter into the Faster-RCNN as their modules share similar functionality.
		We combine the RPN and proposal (re-sample) step from the Faster-RCNN into one and integrate the scoring of both into a single step, resulting in the approach on the right, a two-stage object detector with integrated object permanence.
		As the integration does not change the weights of the model, there is no re-training of an existing detector required.
		Further, leveraging the synergies reduces the computational complexity and is favorable in terms of accuracy as shown by our experiments.
	}
	\label{fig:dpf}
\end{figure*}

\section{Approach}

In this work we present an approach for object permanence in detection which is heavily inspired by particle filters.
Particle filters propagate box candidates between frames, re-sample and score them.
By propagating the box candidates between frames the approach has an explicit model for object permanence.

As our approach requires some understanding of a traditional baseline using Faster-RCNN and a separate particle filter, we will present that first and then describe how to integrate the particle filter into the model to derive our Faster-RCNN detector with object permanence.
All changes to the model will not require any re-training and have minimal computational overhead, when the particle filter is fully integrated.

\subsection{Baseline: Extending Faster-RCNN with Particle Filter} \label{sec:baseline}

First we establish a baseline by extending Faster-RCNN with particle filter as a pre-cursor to our approach.
We choose to use a particle filter, as it shares the conceptual foundation for our approach.
Sharing concepts it is a straightforward stepping stone and allows to clearly attribute all observed performance changes to our contributions.

\textbf{Faster-RCNN:} In our implementation we use a regular Faster-RCNN~\cite{fasterrcnn} re-implementation without any bells and whistles, since it is widely used and a can be considered the gold-standard of two-stage detection.
For example, the MOT Dataset~\cite{mot20} uses Faster-RCNN to provide baseline predictions.

RCNNs are a two stage approach meaning it has a region proposal and a refinement stage.
The region proposal stage uses the entire image to find regions of interest (ROIs) where objects are likely.
The refinement stage then uses ROI crops of the image to refine the proposals by scoring them for the object that is visible and predicting deltas between the proposal and the actual box.
In Faster-RCNN~\cite{fasterrcnn} the first and second stage share an encoder which predicts a feature map and apply the ROI pooling on the feature map greatly reducing the inference time.
Figure~\ref{fig:faster_rcnn} visualizes the architecture.

\textbf{Particle Filter:} A \textit{Particle Filter} predicts the behavior of objects over time using particles.
A particle is a detection with velocity as an additional attribute, that is estimated by the filter.
Our implementation is build from three core components: Resample, measure and predict.

The prediction step in our implementation uses a simple constant velocity linear motion model to predict the position of the particle in the next frame.
We assume a fixed frame rate by using a constant $\Delta t$.

In the measurement step, errors made by the prediction are partially corrected by assigning the particles to detections from the detector and in the case of a successful assignment interpolating linearly.
For this assignment and correction an IoU with the predictions is computed and used to update the scores of particles for subsequent re-sampling step.

The resampling step creates a distribution of particles around interesting regions, compensating uncertainties of the system, \eg motion changes.
To re-sample, we simply use existing particles, remove the particles with the lowest scores and re-sample new particles with a bias towards detections from the Faster-RCNN which have  a low IoU with all particles.
This bias helps to reduce the number of required particles for the filter and improves FPS.

By using a baseline that is consistent with the benchmark defaults and adding a particle filter we can focus on the \textit{integration} of the particle filter into the model and attribute all improvements to the integration. An overview over our baseline architecture is given in Figure~\ref{fig:dpf} on the left.

\subsection{Two-Stage Detection with Integrated Object Permanence (IOP)} \label{sec:dpf}

To integrate object permanence into two stage detection, we merge the particle filter and the Faster-RCNN.
Their components have similar functionalities enabling this integration.
As described in our baseline, a particle filter consists of three steps: Resample, predict and measure.
The goal of the resample step can be described as generating proposals of ROIs and is similar to the RPN in Faster-RCNN.
The measurement step then scores these proposals by their plausibility based on the measurements and can be compared to the scoring and refinement stage in Faster-RCNN.
Finally, the prediction step is used to relate two timesteps and propagate the particles through time, there is no equivalent in two-stage object detectors.

Leveraging the similarities, we merge resampling (proposal generation) and the RPN, as well as the measurement step of the particle filter with the refinement of Faster-RCNN.
The prediction step from particle filter is then used to connect the refinement with the proposal stage.
We call the resulting architecture \textit{Integrated Object Permanence (IOP)} and visualize it in Figure~\ref{fig:dpf} on the right side. However, design decisions when implementing lead to \textit{IOP with particles}, \textit{IOP lite} and \textit{IOP with history}, with varying precision and inference speed.

\begin{figure}
	\centering
	\includegraphics[width=0.48\textwidth]{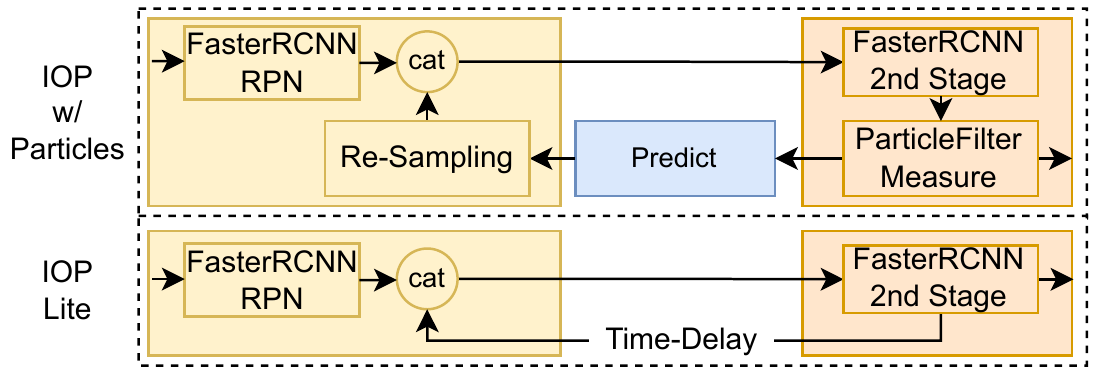}
	\caption{At the core of our implementation for Integrated Object Permanence (IOP) with particles and IOP lite is the depicted information flow. The concatenation of previous predictions and current proposals is a critical component for object permanence in two stage-detection. Measure, predict and re-sampling are skipped in IOP lite.}
	\label{fig:dpf_detailed}
\end{figure}

\textbf{IOP with particles} the most similar to the particle filter baseline.
For resampling, we apply the traditional resample of the particle filter and then use the particle detections as additional proposals for the second stage by concatenating.
Assigning the outputs of the second stage to the particles again is done via IoU based assignment.
This extra step is required, since we use a standard detector without any knowledge of tracklets.
After assignment, the particle filter can be used to predict for the next time frame (Figure~\ref{fig:dpf_detailed}).

\textbf{IOP lite} applies the least changes to an existing two-stage detector, completely omitting tracklets.
The resampling step concatenates the unchanged predictions from the previous frame to the proposals of Faster-RCNN from the current frame, as the RPN is sampling new meaningful proposals.
The measurement step is just the second stage of Faster-RCNN and the prediction step is a simple time-delay (Figure~\ref{fig:dpf_detailed}).

\textbf{IOP with history} is identical to the above except for the resampling step.
Here, predictions from the N previous frames are concatenated to the proposals from the RPN.
One advantage is if an object is fully occluded for a frame and rejected by the second stage, the history allows for quicker recovery once the object is partially visible again.

We trained Faster-RCNN once and all presented variations use the same weights.
There is no training of the individual variations required, as the integration does not change the model itself.
This allows our approach to be integrated into any pre-trained two-stage approach.
Independent of the training data this improves models for inference on sequential data, despite a training on sequential data.

In summary, the core idea of our approach is the feedback of previous predictions as proposals into the model.
Inspired by particle filter, it allows varying degrees of complexity.
Further, keeping weights of the model intact, this idea can be applied to a wide variety of approaches and use cases.

%% file: content/4_results.tex
\section{Experiments}

\input{content/4_results_tab_mot_deltas}
\input{content/4_results_tab_fps}

Our approach is an improvement for object detection.
However, since object permanence is a temporal aspect, we need a tracking dataset for our evaluation.
The evaluation on the tracking dataset will evaluate mainly detection performance gains and computational overhead during inference time.
However, a brief comparison to other trackers on applicable metrics will be done.

For evaluation of our approach for integrated object permanence the challenging MOT17 and MOT20~\cite{mot20} dataset is ideal.
One of the baselines on the dataset is Faster-RCNN which is also part of our baselines: Faster-RCNN with Particle Filter and Faster-RCNN with Kalman Filter.
Comparing against these baselines we will show the effectiveness of the components in our approach.

\subsection{Experimental Setup}

The MOT17 and MOT20~\cite{mot20} dataset consists of 1.423 Frames with 80.274 annotated pedestrians in 11 sequences in total for validation.
The dataset contains heavy occlusions induced by the environment and other pedestrians.
Thus the dataset is very challenging and has multiple situations where object permanence is key for correct detection.

In ablation studies a configuration of Faster-RCNN with a pretrained VGG16 encoder performed best.
No data augmentation or additional datasets were used.
Each sequence was split 80:10:10 and results are reported on the last split which was not used during development and ablation studies.
SGD with momentum (0.9) and a learning rate of 0.001 is used to train the model for 30 epochs.
As an encoder, the model uses VGG16 with batch norm pre-trained on ImageNet~\cite{imagenet}.
For training we follow the standard procedures of Faster-RCNN with sampling 128 samples from the ROIs for training attempting balanced sampling and filling with negative samples.

Kalman Filter still is the gold standard in application.
Thus, we also evaluate a simple Kalman Filter implemented in C++.
This baseline uses above trained Faster-RCNN and no special configuration.

For the particle filter baseline, the same pre-trained Faster-RCNN is used.
The particle filter baseline is optimized for sample efficiency and 200 particles produce sufficiently good results at acceptable inference speed.
Thus we use 50, 75, 100 and 200 particles respectively for our evaluations.

Our approaches, described in Section~\ref{sec:dpf}, use again the same Faster-RCNN. The configuration named \textit{IOP} uses the same number of particles as the particle filter for best comparability.
\textit{IOP lite} has no configuration options and for \textit{IOP with history} we evaluated different history lengths from 1-19 frames, but limit reporting to the best history length of 5 and 9 for IOP and IOP lite respectively in the tracking section.

\subsection{Detection Performance}

As an improvement for detectors, measuring gains in the detection performance is the most important evaluation.
As common in object detection we use the Pascal VOC~\cite{pascalvoc} mean average precision (mAP) metric for evaluation.
We evaluated on a per sequence basis and an average over all sequences (see Table~\ref{tab:mot_results}),
allowing for further insights into the performance.

Averaged over all sequences, our integrated object permanence approach with 200 particles with a gain of +10.3 mAP is best.
IOP lite which is faster and requires no particles achieves an average improvement of +7.32 mAP.
Kalman Filter and Particle Filter can also increase the performance by +4.7 mAP and +7.0 mAP but are outperformed by our IOP or IOP lite.

On MOT20 gains are larger than on MOT17.
It can be observed, that gains are larger on sequences with elevated camera position and lower ego-motion.
In scenes with higher perceived camera motion the gains are smaller and sometimes performance degrades.
This can be explained by the large motion objects have in the image plane and the lack of an ego-motion model to compensate this in all presented approaches.

Overall, our integrated object permanence approach with 200 particle is best in 8 out of 11 sequences.
Our IOP lite is second best in the average category and never degrades the performance of Faster-RCNN, which no other approach was able to achieve.
We explain this by the fact, that IOP lite injects additional proposals and does not remove any proposals or predictions from Faster-RCNN.

\subsection{Latency Overhead}

\input{content/4_results_tab_mota}

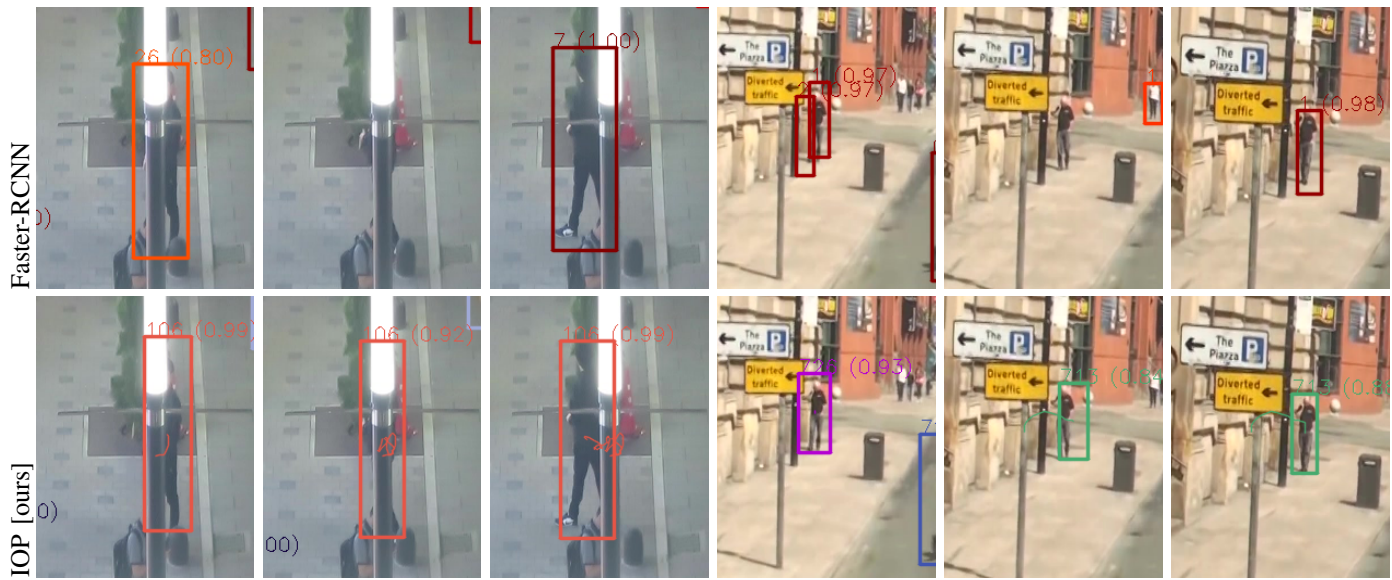
\begin{figure*}
	\centering
	\begin{picture}(6,100)
		\put(0,0){\rotatebox{90}{Faster-RCNN}}
	\end{picture}
	\includegraphics[width=0.158\textwidth]{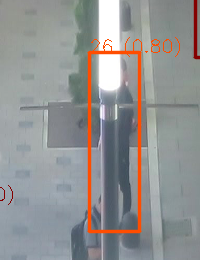}
	\includegraphics[width=0.158\textwidth]{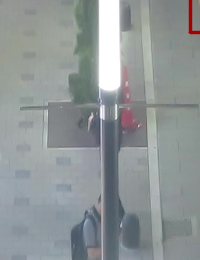}
	\includegraphics[width=0.158\textwidth]{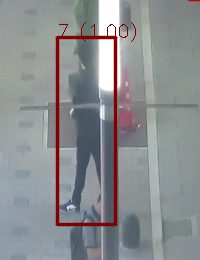}
	\includegraphics[width=0.158\textwidth]{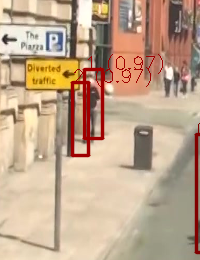}
	\includegraphics[width=0.158\textwidth]{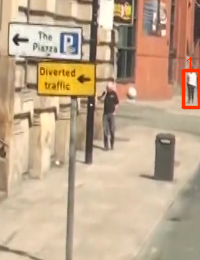}
	\includegraphics[width=0.158\textwidth]{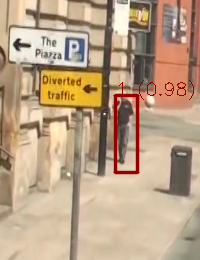}

	\vspace*{0.1cm}

	\begin{picture}(6,100)
		\put(0,0){\rotatebox{90}{IOP [ours]}}
	\end{picture}
	\includegraphics[width=0.158\textwidth]{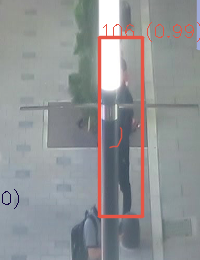}
	\includegraphics[width=0.158\textwidth]{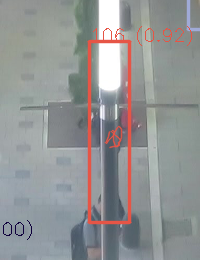}
	\includegraphics[width=0.158\textwidth]{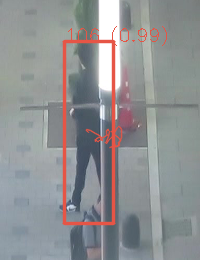}
	\includegraphics[width=0.158\textwidth]{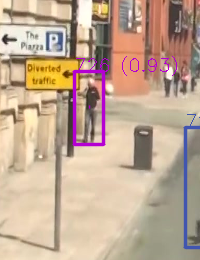}
	\includegraphics[width=0.158\textwidth]{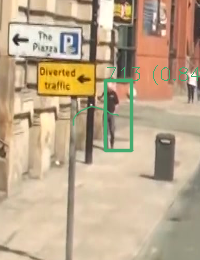}
	\includegraphics[width=0.158\textwidth]{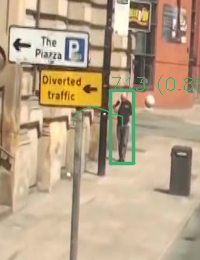}
	\caption{
		Qualitative comparison of Faster-RCNN (top), Integrated Object Permanence (bottom) on three evenly spaced frames.
		The persons in the center of the crops are not detectable by Faster-RCNN for some frames.
		By adding our feedback loop the model is able to correctly detect the person even in sever occlusions (left).
		Similarily, the corner case duplicating a detection and confusing the person for background (right) can be improved despite the ego-motion.
	}
	\label{fig:qualitative_comparison}
\end{figure*}

As latency is an important factor for predictors, we measured the overhead produced by each of the presented approaches compared to the original Faster-RCNN.
We averaged the latency over all samples, to minimize effects caused by the underlying operating systems scheduler.

Our IOP lite (implemented in Python) and Kalman filter (implemented in C++), share the same overhead of 3 ms.
Using particles in a particle filter and our IOP with particles, the overhead is 60-91 ms (see Table~\ref{tab:ablation_studies}). The number of particles has little impact on the overhead.

When speed is more important than the best possible mAP, it is recommended to use IOP lite over IOP with particles, as the overhead is significant.

\subsection{Tracking Metrics}

The main focus is detection.
However, since our approach uses temporal information, a brief evaluation of detection quality focused tracking metrics was done on the MOT17 validation dataset.
For comparison results reported in the publications of the related state-of-the-art approaches were used.
However, in the case of CenterTrack~\cite{zhou2020tracking} FP, FN and IDS are not comparable, since they are reported in percent.

In Table~\ref{tab:mota_comparison} it can be seen that our IOP outperforms Tracktor~\cite{bergmann2019tracking}, CenterTrack~\cite{zhou2020tracking} and FairMOT~\cite{zhang2020fairmot} in all captured metrics except MT.
Our approach is designed to improve detection performance and not primarily as a tracker, however, the integration makes the outputs of the model easily usable by simple IoU based association.

\subsection{Qualitative Analysis}

When qualitatively analyzing the results, we immediately find situations in which the integration of object permanence can change the quality of the output of Faster-RCNN.
In Figure~\ref{fig:qualitative_comparison} example outputs of Faster-RCNN and our approach are visualized.
It can be seen, that on the top some pedestrians cannot be recovered by Faster-RCNN,
but with the feedback loop the pedestrian can be successfully detected with high accuracy.
In the left sequence the person is walking significantly occluded by the pole.
During 28 frames Faster-RCNN detects the person only in 5 frames, with a maximum confidence of 0.5.
In contrast, our proposed IOP detects the person in all frames with a minimum confidence of 0.91.
IOP can stabilize and improve the predictions of a pre-trained Faster-RCNN,
but it is dependent on the Faster-RCNN to detect the object in at least a few frames with low confidence.

%% file: content/4_results_tab_mot_deltas.tex
\begin{table*}[t]
\caption{Comparison of mAP gains on different data sequences. Our Integrated Object Permanence (IOP) with Particles is best overall, whereas our IOP lite never degrades performance at second best overall mAP gains. Experiments use a history length of 1.}
\label{tab:mot_results}
\centering
\begin{tabular}{|c|r|r|rrrr|rrrr|r|}
\hline
\multicolumn{1}{|c|}{\multirow{2}{*}{\textbf{Sequence}}} & \multicolumn{1}{c|}{\multirow{2}{*}{\textbf{FRCNN}}} & \multicolumn{1}{c|}{\multirow{2}{*}{\textbf{\begin{tabular}[c]{@{}c@{}}FRCNN\\ + KF\end{tabular}}}} & \multicolumn{4}{c|}{\textbf{Baseline (FRCNN+ PF)}}                        & \multicolumn{4}{c|}{\textbf{IOP with Particles [ours]}}                  & \multicolumn{1}{c|}{\multirow{2}{*}{\textbf{\begin{tabular}[c]{@{}c@{}}IOP Lite\\~[ours]\end{tabular}}}} \\ \cline{4-11}
\multicolumn{1}{|l|}{}                                   & \multicolumn{1}{l|}{}                                & \multicolumn{1}{l|}{}                                                                               & \textbf{50}      & \textbf{75}      & \textbf{100}     & \textbf{200}     & \textbf{50}      & \textbf{75}      & \textbf{100}     & \textbf{200}     & \multicolumn{1}{l|}{}                                                                             \\ \hline
\hline
MOT17-02  & 39.6 &  +5.6 &   +7.3 &   +7.3 &   +7.3 &   +7.3 &\B{+11.6}&\B{+11.6}&\B{+11.6}&\B{+11.6}&   +5.9 \\ 
MOT17-04  & 78.5 &  +4.0 &   +3.5 &   +4.4 &   +4.4 &   +4.4 &    +6.1 & \B{+6.2}& \B{+6.2}& \B{+6.2}&   +2.5 \\ 
MOT17-05  & 73.3 & -13.3 &\B{+0.3}&\B{+0.3}&\B{+0.3}&\B{+0.3}&    -3.1 &    -3.1 &    -3.1 &    -3.1 &   +0.0 \\ 
MOT17-09  & 73.2 &  +2.5 &   +7.4 &   +7.4 &   +7.4 &   +7.4 & \B{+8.1}& \B{+8.1}& \B{+8.1}& \B{+8.1}&   +4.5 \\ 
MOT17-10  & 26.1 &  +3.6 &   +3.2 &   +3.2 &   +3.2 &   +3.2 & \B{+8.8}& \B{+8.8}& \B{+8.8}& \B{+8.8}&   +7.4 \\ 
MOT17-11  & 51.7 &  -0.4 &\B{+1.5}&\B{+1.5}&\B{+1.5}&\B{+1.5}&    +0.0 &    +0.0 &    +0.0 &    +0.0 &   +0.0 \\ 
MOT17-13  & 12.7 &  +2.5 &   -2.4 &   -2.4 &   -2.4 &   -2.4 &    +7.3 &    +7.3 &    +7.3 &    +7.3 &\B{+7.9}\\ \hline
MOT20-01  & 66.6 &  +7.4 &  +12.6 &  +12.6 &  +12.6 &  +12.6 &\B{+16.1}&\B{+16.1}&\B{+16.1}&\B{+16.1}&  +12.1 \\ 
MOT20-02  & 72.3 &  +5.3 &   +6.9 &   +8.3 &   +8.3 &   +8.3 &\B{+11.0}&\B{+11.0}&\B{+11.0}&\B{+11.0}&   +5.5 \\ 
MOT20-03  & 41.9 & +19.9 &  -10.7 &   +3.2 &  +12.6 &  +20.2 &    -8.1 &    +5.6 &   +18.4 &\B{+28.3}&  +22.2 \\ 
MOT20-05  & 55.4 & +14.3 &  -25.2 &  -11.2 &   -0.4 &  +14.5 &   -24.3 &    -7.7 &    -0.7 &\B{+18.7}&  +12.6 \\ \hline \hline
MINIMUM   & 12.7 & -13.3 &  -25.2 &  -11.2 &   -2.4 &   -2.4 &   -24.3 &    -7.7 &    -3.1 &    -3.1 &\B{+0.0}\\
MAXIMUM   & 78.5 & +19.9 &  +12.6 &  +12.6 &  +12.6 &  +20.2 &   +16.1 &   +16.1 &   +18.4 &\B{+28.3}&  +22.2 \\
AVERAGE   & 53.8 &  +4.7 &   +0.4 &   +3.2 &   +5.0 &   +7.0 &    +3.0 &    +5.8 &    +7.6 &\B{+10.3}&   +7.3 \\ \hline
\end{tabular}
\end{table*}

%% file: content/4_results_tab_fps.tex
\begin{table}[t]
\renewcommand{\arraystretch}{1.0}
\caption{Average latency overhead of the tested approaches on MOT.}
\label{tab:ablation_studies}
\centering
\begin{tabular}{|l|r|r|r|}
\hline
\textbf{Approach}    & \textbf{\#Particles} & \textbf{mAP} & \textbf{Latency} \\ \hline
\hline
Faster-RCNN                                &   - &    53.8 &  550 ms \\ \hline
Kalman Filter                              &   - &    +4.7 &\B{+3 ms}\\ \hline
\multirow{4}{*}{Particle Filter}           &  50 &    +0.4 &  +76 ms \\
                                           &  75 &    +3.2 &  +57 ms \\ 
                                           & 100 &    +5.0 &  +60 ms \\ 
                                           & 200 &    +7.0 &  +63 ms \\ \hline
\multirow{4}{*}{IOP with Particles [ours]} &  50 &    +3.0 &  +91 ms \\
                                           &  75 &    +5.8 &  +74 ms \\
                                           & 100 &    +7.6 &  +79 ms \\
                                           & 200 &\B{+10.3}&  +79 ms \\ \hline
IOP Lite [ours]                            &   - &    +7.3 &\B{+3 ms}\\ \hline
\end{tabular}
\end{table}


%% file: content/4_results_tab_mota.tex
\begin{table*}[t]
\renewcommand{\arraystretch}{1.0}
\caption{
	Evaluating tracking metrics on MOT 17 validation set, the overall best approach is IOP lite with a history length of 9 frames.
	\\
	$^1$Numbers in percent are excluded from comparison.
	$^2$Numbers not reported in original paper and not on validation split.
}
\label{tab:mota_comparison}
\centering
\begin{tabular}{|l|c|c|c|c|c|c|c|c|c|c|}
	\hline
	\textbf{Approach} & \textbf{Hist} & \textbf{MOTA}   & \textbf{MOTP}   & \textbf{IDF1}   & \textbf{MT}   & \textbf{ML} & \textbf{FP} & \textbf{FN} & \textbf{IDS} & \textbf{DetA} \\
	\hline 	\hline
	Tracktor~\cite{bergmann2019tracking}   & - & 61.9 & - & 64.7 & 35.3 & 21.4 & 323     & 42454   & 326    & - \\
	CenterTrack~\cite{zhou2020tracking}    & - & 66.1 & - & 64.2 & 41.3    & 21.2 & 4.5\%$^1$  & 28.4\%$^1$ & 1.0\%$^1$ & - \\
	FairMOT~\cite{zhang2020fairmot}        & - & 69.1 & - & 72.8 & -             & -    & -       & -       & 299    & 42.9$^2$	 \\
	PermaTrack~\cite{tokmakov2021learning} & - & 69.5 & - & 71.9 & 41.0          & 39.5 & -       & -       & -      & 58.0$^2$ \\
	\hline
	IOP w/ Particles [ours] & 1 & 52.3          & 34.1          & 67.8          & \textbf{123} & \textbf{12} & 1591        & \textbf{678} & 231         & 70.6          \\
	IOP w/ Particles [ours] & 5 & 52.5          & 31.4          & 69.1          & {\ul 114} & \textbf{12} & 1554        & {\ul 750}    & 200         & 70.7          \\
	IOP lite [ours]         & 1 & {\ul 69.4}    & {\ul 61.7}    & {\ul 75.3}    & 89  & 37          & {\ul 131}   & 1817         & \textbf{46} & {\ul 70.8}    \\
	IOP lite [ours]         & 9 & \textbf{72.5} & \textbf{63.5} & \textbf{77.1} & 93  & 34          & \textbf{83} & 1685         & {\ul 48}    & \textbf{73.7} \\
	\hline
\end{tabular}
\end{table*}

%% file: content/5_conclusions.tex
\section{Conclusions}

Object detection is applied to safety-critical domains, \eg robotics and autonomous vehicles.
However, current detectors lack the concept of object permanence, leading to temporally unstable predictions, \eg with temporary occlusion.
Trackers solve this but are complex or need special training data.

Our Integrated Object Permanence (IOP) fills this gap by introducing object permanence into two-stage approaches without the need for re-training.
Inspired by a particle filter, a feedback loop is integrated into a Faster-RCNN.
At the core, predictions of previous frames are used as proposals for the next frame.

In multiple experimental setups, the effects of each design decision are evaluated and we conclude, that for most use-cases IOP lite is the best option, as it has best or second to best performance with least computational overhead.
IOP with particles is best in object detection, but IOP lite is best in inference speed and tracking on most metrics.

Our approach is an ideal solution to improve object detection performance without any need for sequential training data.
As no retraining is needed, we can apply this approach to already existing two-stage detectors, like Faster-RCNN.
The concept is general and only requires a proposal and refinement step in the model and has little computational overhead as exhibited by IOP lite.
Thus, we see a wide range of applications and use-cases where IOP can improve object detection.
For example in instance segmentation or even human pose estimation.

%% file: content/6_acknowledgment.tex
\section*{ACKNOWLEDGMENT}


The research leading to these results is funded by the German Federal Ministry for Economic Affairs and Energy within the project ”KI-Absicherung” (grant: 19A19005U)
and the German Federal Ministry of Education and Research within the project "VIZTA" (grant: 16ESE0424 / GA826600).